\documentclass[a4paper,fleqn]{cas-dc}

\usepackage[authoryear,longnamesfirst]{natbib}

\definecolor{mypink2}{RGB}{219, 48, 122}

\begin{document}
\let\WriteBookmarks\relax
\def\floatpagepagefraction{1}
\def\textpagefraction{.001}

\shorttitle{Detection-Guided Deep Learning-Based Model with Spatial Regularization for Lung Nodule Segmentation}

\shortauthors{Jiasen Zhang et~al.}

\title [mode = title]{Detection-Guided Deep Learning-Based Model with Spatial Regularization for Lung Nodule Segmentation}                      

\author[1,2]{Jiasen Zhang}

\author[2,3]{Mingrui Yang}
\author[1]{Weihong Guo}
\author[4]{Brian A. Xavier}
\author[4]{Michael Bolen}
\author[2,3,5]{Xiaojuan Li}
\cormark[1]

\affiliation[1]{organization={Department of Mathematics, Applied Mathematics, and Statistics, Case Western Reserve University},
    addressline={2145 Adelbert Rd}, 
    city={Cleveland},
    state={OH},
    postcode={44106}, 
    country={United States}}
\affiliation[2]{organization={Program for Advanced Musculoskeletal Imaging (PAMI), Cleveland Clinic},
    addressline={9620 Carnegie Ave N Bldg}, 
    city={Cleveland},
    postcode={44106}, 
    state={OH},
    country={United States}}
\affiliation[3]{organization={Department of Biomedical Engineering, Lerner Research Institute, Cleveland Clinic},
    addressline={9620 Carnegie Ave N Bldg}, 
    city={Cleveland},
    postcode={44106}, 
    state={OH},
    country={United States}}
\affiliation[4]{organization={Imaging Institute, Cleveland Clinic},
    addressline={9500 Euclid Ave}, 
    city={Cleveland},
    postcode={44195}, 
    state={OH},
    country={United States}}
\affiliation[5]{organization={Department of Biomedical Engineering, Case Western Reserve University},
    addressline={10900 Euclid Ave}, 
    city={Cleveland},
    state={OH},
    postcode={44106}, 
    country={United States}}

\cortext[cor1]{Corresponding authors email addresses: lix6@ccf.org}



\begin{abstract}
Lung cancer ranks as one of the leading causes of cancer diagnosis and is the foremost cause of cancer-related mortality worldwide. The early detection of lung nodules plays a pivotal role in improving outcomes for patients, as it enables timely and effective treatment interventions. The segmentation of lung nodules plays a critical role in aiding physicians in distinguishing between malignant and benign lesions. However, this task remains challenging due to the substantial variation in the shapes and sizes of lung nodules, and their frequent proximity to lung tissues, which complicates clear delineation. In this study, we introduce a novel model for segmenting lung nodules in computed tomography (CT) images, leveraging a deep learning framework that integrates segmentation and classification processes. This model is distinguished by its use of feature combination blocks, which facilitate the sharing of information between the segmentation and classification components. Additionally, we employ the classification outcomes as priors to refine the size estimation of the predicted nodules, integrating these with a spatial regularization technique to enhance precision. Furthermore, recognizing the challenges posed by limited training datasets, we have developed an optimal transfer learning strategy that freezes certain layers to further improve performance. The results show that our proposed model can capture the target nodules more accurately compared to other commonly used models. By applying transfer learning, the performance can be further improved, achieving a sensitivity score of 0.885 and a Dice score of 0.814.
\end{abstract}




\maketitle

\section{Introduction}
Lung cancer has been one of the most deadly forms of cancer throughout the 20th century. In 2023, it accounted for the second-highest number of new cancer cases and remained the leading cause of cancer-related deaths \cite{cancerStat}. Early-stage lung cancer often presents with no clear symptoms, making lung nodules critical markers for identifying the disease. Consequently, the early detection of lung nodules is paramount for the prompt diagnosis and treatment of lung cancer, significantly enhancing the chances of patient survival. Computed tomography (CT) imaging plays a pivotal role in the early detection of lung tumors, providing radiologists with the ability to identify and examine suspicious lesions at their most treatable stage. Typically, CT scans are initially reviewed by radiologists to check for the presence of nodules, which are then categorized as benign or malignant based on their size and shape, or in some instances based upon appearance over time. This process, however, can be time-consuming and labor-intensive, often involving the review of hundreds or even thousands of image slices per scan, and may lead to varying interpretations among different radiologists. Image segmentation techniques offer a solution by precisely delineating nodules from the surrounding lung tissue, offering valuable data on their size, shape, and growth rate. Therefore, the development of accurate and automated lung nodule segmentation methods is crucial. These technologies not only assist radiologists in the identification and classification of nodules but also significantly reduce their workload, and have the potential to increase efficiency, leading to improved diagnostic accuracy and patient care.

In recent decades, various medical image segmentation methods have been developed, enhancing the precision of medical diagnoses. Thresholding techniques, one of the earliest methods, segregate an image into two parts using a predefined intensity threshold \cite{threshold1}. A more advanced approach has evolved to allow segmentation into multiple sub-images, each with its tailored thresholds for more nuanced analysis \cite{threshold2}. Morphology methods \cite{morpho} leverage predefined shapes, scanning images to identify areas that closely match these criteria for segmentation. Variational techniques, on the other hand, minimize specific energy functions to ensure the segmented boundaries align accurately with the image's target structures, exemplified by the Mumford–Shah model, which segments images into uniform intensity regions \cite{mumford}. Further sophistication is seen in methods employing level set functions, where boundaries are represented mathematically, and energy functions are crafted to minimize within-region heterogeneity. The Chan-Vese method is a special case of Mumford-Shah model that uses level sets to segment images that do not necessarily have clear edges \cite{chanvese1, chanvese2}. Graph cut methods represent an image as a weighted undirected graph and employ algorithms to enhance sub-graph similarity while distinguishing between different sub-graphs \cite{graphcut1,graphcut2}. Recent advancements also include unsupervised machine learning techniques like K-means \cite{kmeans}, KNN (k-Nearest Neighbor) \cite{knn}, and probabilistic models \cite{probabilistic}, further broadening the toolkit available for medical image segmentation.

With the advancement of computer technology, deep learning models, particularly convolutional neural networks (CNNs) \cite{cnn} have surpassed traditional methods in efficacy across numerous computer vision tasks. Traditional techniques often depend on manual feature selection and prior knowledge, whereas deep learning models autonomously learn to extract relevant features at various levels directly from data. Among these, U-Net \cite{unet} has become a benchmark model in medical image segmentation for its efficiency and simplicity. This model utilizes an encoder-decoder framework with added skip connections, allowing for the detailed capture and preservation of spatial information, a crucial aspect in accurately delineating medical images. The encoder reduces feature dimensions to capture hierarchical information, while the decoder restores these features to their original resolution, ensuring precise segmentation. The skip connections help preserve spatial information and aid in the precise localization of structures, which is crucial in medical image segmentation tasks. The introduction of U-Net variants, such as ResUNet \cite{resunet} and ResUNet++ \cite{resunetplus}, further enhances segmentation accuracy. Concurrently, semantic segmentation models like DeepLab \cite{deeplab, deeplabv2, deeplabv3, deeplabv3plus}  have set new standards on public datasets for semantic segmentation like PASCAL VOC 2012 \cite{pascal-voc-2012} by incorporating advanced features such as Xception \cite{xception} and atrous spatial pyramid pooling (ASPP) \cite{deeplabv2}, showing strong potential for application in medical images. These developments underscore the potential of deep learning models to revolutionize medical image analysis, offering significant improvements in diagnostics and treatment planning.

\begin{figure*}[ht]
\centering
\includegraphics[width=1\textwidth]{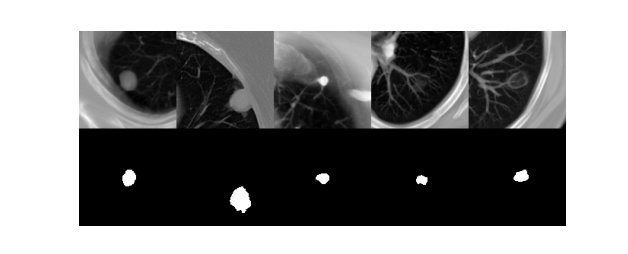}
\caption{Examples of nodules from LUNA16. From left to right: isolated nodule, juxtapleural nodule, calcific nodule, ground-glass opacity (GGO) nodule and cavitary nodule.}
\label{figure1}
\end{figure*}

While numerous lung nodule segmentation methods have been introduced, the task remains daunting. A primary challenge is the limited size of clinical datasets for lung nodule segmentation, as nodules are infrequent compared to the prevalence of normal lung tissue in imaging studies. Additionally, lung nodules exhibit a wide variety in shape, size, and attenuation, complicating their classification and segmentation. Generally, lung nodules can be classified into several types: isolated nodules, juxtapleural nodules, cavitary nodules, ground-glass opacity (GGO) nodules, and calcific nodules \cite{type}. As depicted in Figure~\ref{figure1}, juxtapleural nodules are connected to the lung pleura and chest wall and exhibit similar attenuation to it. Cavitary nodules pose challenges for accurate segmentation due to variations in attenuation in setting of central lucency. Calcific and GGO nodules display high and low attenuation values, respectively. Moreover, the presence of numerous small nodules, which can fall into any of these categories, adds another layer of complexity to achieving accurate segmentation. Given the predominance of small size of nodules within clinical datasets relative to the size of CT scans, our approach concentrates on segmenting small regions of interest (ROI) around these nodules. Focusing on ROIs, identified either by radiologists or automated systems, allows for the prioritization of suspect nodules for in-depth analysis and clinical decision-making. It is usually integrated with Computer-aided detection (CAD) systems, which allows for streamlined workflows, and enables more precise delineation of nodules and more efficient utilization of computational resources in automated nodule detection and analysis. Such targeted efficiency is vital in clinical environments where swift diagnostic and treatment decision-making are paramount for radiologists.

In the field of deep learning, several strategies have been developed to overcome segmentation challenges and address the scarcity of training data. Recent innovations include applying transfer learning from extensive datasets \cite{transferlearning1, transferlearning2}, multitask learning to leverage shared features across tasks like classification and segmentation \cite{multitask-jsc}, and employing data augmentation, cross-validation, and regularization to mitigate overfitting.

This paper presents a multitask deep learning model that unites segmentation and classification tasks, leveraging shared information through feature combination blocks. The model undergoes pre-training on the large-scale public LUNA16 dataset before adapting to a smaller, private clinical dataset through an optimal transfer learning strategy and enhancement by spatial regularization. The key contributions of this paper include:

\begin{enumerate}
\itemsep=0pt
\item We propose a multitask model that enhances segmentation performance through integrated classification features.
\item We introduce a feature combination block for efficient, flexible feature utilization without information loss, avoiding unnecessary upsampling and downsampling. 
\item We apply soft threshold dynamic (STD) regularization for refined segmentation predictions, incorporating classification outputs as prior information for enhanced segmentation accuracy. 
\item We adopt an optimal transfer learning strategy facilitating effective use of pre-training data for segmentation on a smaller clinical dataset. 
\item Our results demonstrate superiority of our model over other classic medical image segmentation models, supported by experimental evidence of improved performance through feature combination, spatial regularization, and transfer learning. 
\end{enumerate}  

The remaining part of this paper is organized as follows: Section 2 discusses related literature studies; Section 3 explains the proposed methodology; Section 4 presents our experimental results, with discussions provided for each experiment; and finally, Section 5 presents the conclusion of the study.

\section{Related works}
\subsection{Feature combination in multitask model}
Multitask learning is a technique in which a model is trained to perform multiple tasks simultaneously. By leveraging information from related tasks, multitask learning can improve performance and generalization ability, which is especially helpful when training data is limited. Additionally, multitask learning can serve as a form of regularization by introducing additional constraints on the model.

When training a multitask model, it is essential to combine the features of different tasks to allow them to share information; otherwise, the tasks would operate independently. The feature combination, or information fusion scheme, usually depends on the architecture of deep neural networks. Conventional multitask models with a single channel \cite{multitask-effynet, multitask1} typically share parameters across different tasks in the hidden layers. The final outputs include predictions for different tasks, which may be generated in the middle or final layers of the model. However, tasks in computer vision, such as segmentation, detection, and classification, often emphasize different types of features. Sharing too many parameters could limit the model's performance across tasks. To address this issue, some models \cite{multitask-hfnet} split the single channel into double channels for two tasks or use separate architectures \cite{multitask-jsc}. Misra et al. \cite{multitask2} evaluated the performance of two-task architectures split at different layers and concluded that the optimal multitask architecture depends on the individual tasks. It is essential to design feature combination blocks for split architecture so that the multiple tasks can share information and improve each other, otherwise the different tasks will just work separately. Pure concatenation and linear combination \cite{multitask2} were naturally used and shown positive effects. A more general way is to apply more hidden layers in feature combination blocks. The model in \cite{multitask-jsc} upsampled the smaller feature so that the features have the same size and can be concatenated together. However, like downsampling, upsampling could also cause loss of information because it involves replicating or interpolating existing information to fill in the gaps.

\subsection{Soft Threshold Dynamic (STD) regularization}
To combine the deep learning framework with spatial regularization, Liu et al. \cite{liujun1, liujun2, liujun3} provided variational explanations for some widely used activation functions in deep learning, including softmax, ReLU, and sigmoid. For instance, the sigmoid function can be expressed as: $Sig(u) = \frac{1}{1 + \exp(-u)}$ and it can be represented as the argument that solves the following convex optimization problem when the parameter $\varepsilon$ is equal to 1:
\begin{equation}
\begin{split}
Sig(u) = \mathop{\arg\min}_{0\le x \le 1} -\langle u, x\rangle + \varepsilon\langle x,\ln x \rangle + \varepsilon\langle(1-x), \ln(1-x) \rangle
\label{sfx}
\end{split}
\end{equation}
where $u$ is the feature map input, and $ \langle a, b \rangle$ means inner product of $a$ and $b$. The optimal solution  $x$ will be equal to $Sig(u/\varepsilon)$, which can be proved by simply computing the derivative of the energy function. One can easily add regularization terms used in image processing and computer vision into it to achieve smoothness. In \cite{liujun3}, soft threshold dynamics (STD) \cite{td} regularization was adopted to control the length of the boundary so the edges have certain smoothness instead of being bumpy:
\begin{equation}
\begin{split}
Sig_{STD}(u) = & \mathop{\arg\min}_{0\le x \le 1} -\langle u, x\rangle + \varepsilon\langle x,\ln x \rangle \\
& + \varepsilon\langle(1-x), \ln(1-x) \rangle \\
& + \lambda\langle x, k_\sigma *(1-x)\rangle 
\label{std}
\end{split}
\end{equation}
where $*$ means convolution, $k$ represents a discrete Gaussian kernel with standard deviation $\sigma$ and $\lambda$ is a parameter. The STD regularization term $\langle x, k_\sigma *(1-x)\rangle$ has been proved to approximate the boundary length \cite{mine} and has similar effect as total variation (TV) regularization. It involves convolution and inner product operation, which are cheaper to compute. Compared to Total Variation (TV) regularization, which is non-smooth, STD regularization is more efficient in deep learning due to its smooth nature, making it compatible with neural networks and easier to solve. Furthermore, the new term (last term) is concave so that problem~(\ref{std}) can be solved with an efficient and iterative algorithms. Given the input $u$ which is the output of the last hidden layer, the initial value $x^0$ can be set as $x^0 = Sig(u)$. Then for each t-th step, the value of the next step can be computed as:
\begin{equation}
\begin{split}
x^{t+1} = Sig \Bigg(  \frac{x^t - \lambda k_\sigma * (1-2x^t)}{\varepsilon} \Bigg)
\label{stdsolve}
\end{split}
\end{equation}

It is important to note that the network with STD regularization differs from post-processing methods like Conditional Random Fields (CRF) because it is integrated into the back-propagation process and influences gradient updating. In practice, parameters such as $\varepsilon$, $\lambda$, and $\sigma$ can be made learnable, allowing them to be automatically tuned by the neural network.

\section{Proposed methodology}

\subsection{Model architecture}
\begin{figure*}[ht]
\centering
\includegraphics[width=0.9\textwidth]{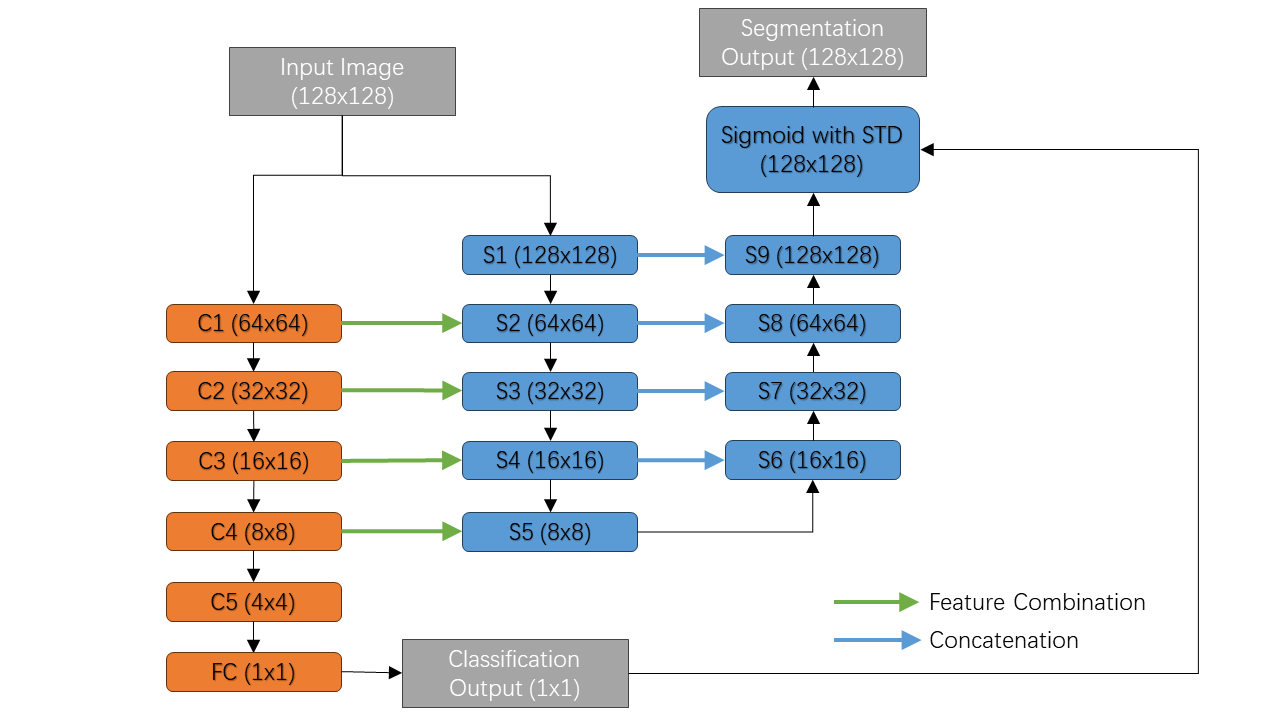}
\caption{The architecture of the proposed multitask model. The orange blocks correspond to ResNet-50 for classification.  The blue blocks correspond to an ResU-Net based network for segmentation. An input of size $128\times 128$ is used for demonstration.}
\label{figure2}
\end{figure*}

The proposed multitask model aims to leverage the information of classification to improve the performance of segmentation. The architecture is illustrated in Figure~\ref{figure2}, and the size of the output of each block is also shown. The orange blocks correspond to a ResNet-50 binary classifier whose final activation function is Sigmoid. In our model, the classifier learns to output a score between zero and one, representing the probability of the existence of a nodule. Because the classification task is relatively simpler than segmentation, we choose ResNet-50 to guarantee efficiency without losing accuracy. The blue blocks correspond to a ResU-Net based segmentation model. The original ResU-Net model \cite{resunet} consists of an encoder (blocks S1 to S5) and a decoder (blocks S6 to S9). In the encoder part, the size of the features gets smaller gradually and the number of channels gets larger to capture some high-level features. This means the encoder part works similarly to ResNet-50, and therefore it is appropriate to combine the features of ResNet-50 and the U-Net encoder. The feature combination blocks are represented by green arrows in Figure~\ref{figure2}, which will be introduced later. Finally, after getting the output of the classification, we embed it into the sigmoid layer before the segmentation output. Different from feature combination, we expect the classification output to directly affect the segmentation output.

In the model, we modified the ResU-Net to achieve better performance. Originally, each ResU-Net block includes two BRC blocks (a batch normalization layer, a ReLU, and a convolution layer). The first convolution layer has a stride of 2 to reduce the size of features. We replaced it with a depth-wise separable convolutional layer used in Xception modules \cite{xception} to reduce computational cost and improve performance. This decomposition involves splitting the standard $3\times 3$ convolutional layer into a $3\times 3$ channel-wise spatial convolution followed by a $1\times 1$ point-wise convolution. Additionally, inspired by ResUNet++ \cite{resunetplus}, we added an atrous spatial pyramid pooling (ASPP) \cite{deeplabv2} module respectively at the end of the encoder and decoder (S5 and S9), which can also significantly improve performance. ASPP typically consists of four atrous convolutions with different dilation rates to help increase the receptive field and capture multi-scale information. It is suggested that the dilation rates should not have common factor relationships like 2, 4, 8, etc. \cite{dilation}, so we chose dilation rates of 1, 5, 10, and 15.

\subsection{Feature combination block}
\begin{figure}[ht]
\centering
\includegraphics[width=0.5\textwidth]{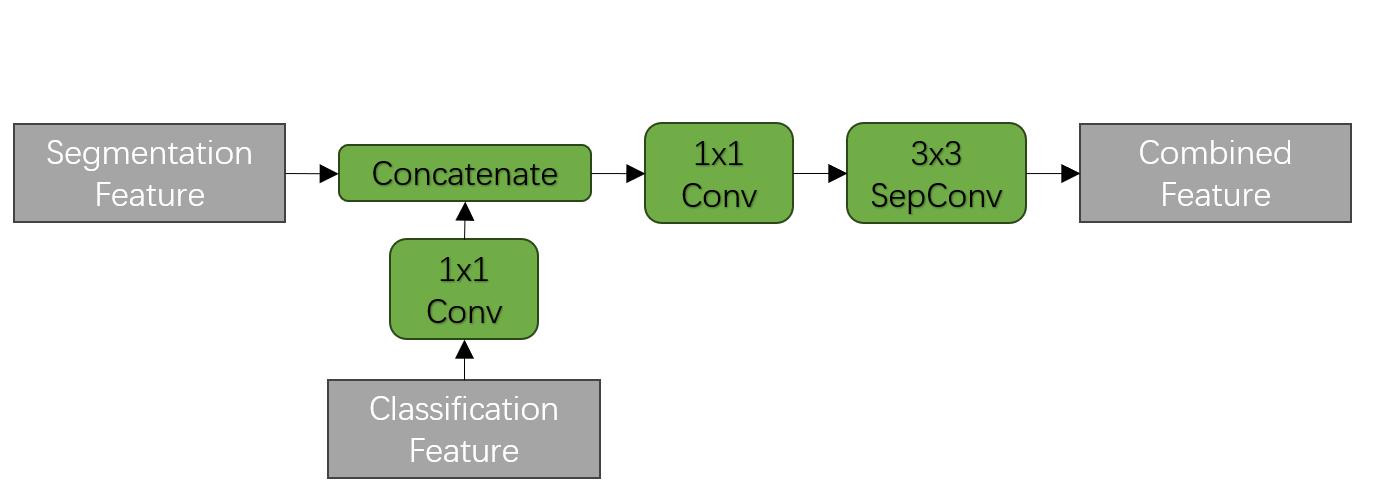}
\caption{Structure of our combination block.}
\label{figure3}
\end{figure}

Since we expect the multitask model to leverage the features of the classification model in order to improve the performance of the segmentation model, it is necessary to design an appropriate scheme to merge the classification and segmentation features. In our method, we only combine the features with the same size so that there is no up-sampling and down-sampling. Therefore, the combination strategy is flexible and depends on the specific model architecture. The combined blocks are indicated in Figure~\ref{figure3}, in which the final layers are used as the combined feature. Instead of concatenating two features directly, we first let the classification feature go through a $1\times 1$ convolutional layer (the number of output channels is equal to that of the input channel) so that it is transformed into a segmentation feature. Then the two features are concatenated together. We apply another $1\times 1$ convolutional layer so that the number of channels is reduced to the number of segmentation channels. Because the number of channels of the input features remains unchanged, the second $1\times 1$ convolutional layer will learn to fuse the concatenated features together. Finally, the combined feature is fed into a $3\times 3$ depth-wise separable convolutional layer with the same number of input and output channels, which is the number of segmentation channels.

\subsection{Embedded STD regularization}
Apart from feature combination, we utilize the result of classification to improve segmentation. We incorporate it as a prior combined with STD regularization, and the new optimization problem can be expressed as in Equation~\ref{newstd}. Suppose the classification output is denoted as $c$ which is between zero and one, representing the probability that the patch contains nodule(s). When $c = 0$ there is no nodule in the patch and when $c=1$ there is at least one nodule. We add a new term $(1-c) \langle I, x \rangle$ to the STD regularization. Note $\langle I, x \rangle$ measures the size/volume of the nodule. When $c$ is close to 0, the classification indicates that there is low probability of having nodules and the additional term controls the size of the nodules.  On the other hand, when $c$ is close to or equal to one, the classifier is confident that there is something resembling a nodule. In this case, $1-c$ is small, and the new prior has almost no effect.  Thus, the new term has the effect of reducing the size of the nodule in the prediction if the classification indicates a high chance of noting having nodules ($c$ being close to or equal to 0). This is very helpful in reducing false positives and avoiding identifying other tissues as nodules. It effectively combines classification output into segmentation to help improve segmentation results.  In the neural network, the $c$ in Equation~\ref{newstd} is excluded from backpropagation to avoid affecting the classification result. 
\begin{equation}
\begin{split}
Sig_{prop}(u) = & \mathop{\arg\min}_{0\le x \le 1} -\langle u, x\rangle + \varepsilon\langle x,\ln x \rangle + \\
& \varepsilon\langle(1-x), \ln(1-x) \rangle \\
& + \lambda_1 \langle x, k_\sigma *(1-x)\rangle \\
& + \lambda_2(1-c) \langle I, x \rangle
\label{newstd}
\end{split}
\end{equation}

Both $\lambda_1$ and $\sigma$ control the effect of smoothness. When we use STD regularization in our models, we set $\lambda_1$ to be equal to one and tune $\lambda_2$ and $\sigma$ automatically. Since the new term $(1-c) \langle I, x \rangle$ and its derivative are constant, the solution of Equation~\ref{newstd} is efficient and similar to that of Equation~\ref{std}. The new iterative scheme is shown as in Equation~\ref{newstdsolve}. The initial value could be set as $Sigmoid(u)$, and $x$ will converge quickly within ten steps.
\begin{equation}
\begin{split}
x^{t+1} = Sig \Bigg(  \frac{x^t - \lambda_1 k_\sigma * (1-2x^t) - \lambda_2(1-c)}{\varepsilon} \Bigg)
\label{newstdsolve}
\end{split}
\end{equation}

\section{Experiments and results}
\subsection{Dataset}
Two kinds of datasets are used in our experimental evaluation: one for pre-training, and the other one for transfer training and performance evaluation. Because annotated clinical lung nodule dataset is typically limited in size, we use the Lung Nodule Analysis 2016 (LUNA16) dataset \cite{luna16} for pre-training. This dataset is a subset of the LIDC-IDRI dataset \cite{lidc}, containing 1018 CT scans annotated by four radiologists. In LUNA16, scans thicker than 3mm and those with inconsistent slice spacing or missing slices were removed, resulting in a total of 888 scans \cite{luna16}. For each scan, multiple annotations were merged so that their positions and diameters were averaged. Our method focuses on segmenting lung nodules within a predefined region of interest (ROI) containing lung nodules. In total, we obtained 2523 nodules. For each nodule, we found the location of the centroid and then crop a 3D region centered around the centroid so that each cropped region contains only one nodule. The first two dimensions of the 3D region is 128 by 128 and the third dimension depends on the lesion size. Note the proposed method is based on two-dimensional segmentation and some 2D slices without nodule are remained for pre-training. 

We evaluated the performance of the proposed method on a small lung nodule dataset from Cleveland Clinic (CCF). It is collected from 7 patients and annotated by one radiologist. Similarly, 3D regions (128 by 128 along the first two dimensions) were cropped around each lesion without overlap. In total, the dataset from 7 patients consists of 56 lesions, 45 of which were randomly selected and used for training and 5-fold cross-validation. The remaining 11 lesions were used as a testing set for performance evaluation. In the CCF dataset, each CT scan typically contains hundreds of slices, while lung nodules only appear in about ten of the slices. Since the proposed model is 2D-based and has a classification component, most slices without nodules are excluded from the dataset but we do keep some slices that have no nodules to train the classifier. Some empty slices that are adjacent to the lung nodule are retained so that the ratio of slices with and without nodules is balanced. This resulted in a total of 814 slices remaining after the clean-up.

The image intensity of the LUNA16 dataset ranged from 0 to 255, while the intensity of the CCF dataset ranged from 0 to 1000. To make the pre-trained model compatible for transfer learning, both datasets were normalized according to the equation below so that their intensities are in the range of 0 to 1. Additionally, we applied random flips in the data augmentation process.

\begin{align}
I_{normalize} = \frac{I - I_{min}}{I_{max} - I_{min}}
\end{align}

\subsection{Implementation details}
First, we pre-trained the model with the LUNA16 dataset for 200 epochs with a batch size of 10. The initial learning rate was set to 0.001 without decay. Then, we trained the model with the CCF dataset for 50 epochs with a batch size of 10, which was sufficient considering its small size. The initial learning rate was 0.001 with a decay of 0.75 for every five epochs. We used Adam as the optimizer with a weight decay of 1e-8. The loss function for segmentation was a combination of binary cross-entropy (BCE) and Dice loss. The binary classification also used BCE loss. During model training, the overall loss was calculated as follows:
\begin{align}
L = L_{Dice} + \lambda_1 L_{BCE}^{seg} + \lambda_2 L_{BCE}^{class}
\end{align}
In our experiments both $\lambda_1$ and $\lambda_2$ were set to be one.

To utilize the information learned from LUNA16, we conducted a series of experiments to determine the most effective transfer learning strategy. Our conclusion was that the best results are achieved when the model is fully fine-tuned, meaning no layers are frozen. Additionally, STD regularization was not applied during pre-training.

\subsection{Evaluation metrics}
To ensure a comprehensive evaluation of the models' performance, we employ several commonly used metrics: precision (positive predictive value or PPV), sensitivity (or recall), Dice score (also called F1 score), and IoU (intersection over union). In the case of binary segmentation, they are defined as follows:
\begin{align}
& \textrm{Precision} = \frac{\textrm{TP}}{\textrm{TP}+\textrm{FP}} \\
& \textrm{Sensitivity} = \frac{\textrm{TP}}{\textrm{TP}+\textrm{FN}} \\
& \textrm{Dice} = \frac{2*\textrm{Sen}*\textrm{Pre}}{\textrm{Sen} + \textrm{Pre}} = \frac{2\textrm{TP}}{2\textrm{TP} + \textrm{FN} + \textrm{FP}} \\
& \textrm{IoU} = \frac{\textrm{TP}}{\textrm{TP} + \textrm{FN} + \textrm{FP}} 
\end{align}
where TP represents true positive, i.e., the number of truly classified pixels whose true values are positive. Similarly, TN, FP, and FN are true negative, false positive, and false negative. Suppose the truth value of lung nodule area is positive and that of non-infected is negative, precision measures the accuracy rate of nodules in terms of prediction. Sensitivity measures the accuracy rate of nodules in terms of ground truth. Dice score is a combination of precision and sensitivity, measuring the similarity between the predicted lung infections and the ground truth. IoU is another metric measuring the similarity of two finite sets, defined as their intersection divided by their union. In some cases IoU is also related to volumetric overlap error (VOE) by $\text{IoU} = 1- \text{VOE}$.

Another widely used evaluation metric in medical image segmentation is Hausdorff distance (HD), measuring the maximum distance from points in one contour to the closest point in the other contour. It provides a straightforward metric for comparison between the segmented region and the ground truth. Suppose the contours of the prediction and ground truth domains are $X$ and $Y$, their Hausdorff distance is defined as
\begin{align}
\textrm{HD} = \max \{ \sup_{x\in X} d(x, Y), \sup_{y\in Y} d(X, y) \}
\end{align}
where $d(x, Y)$ is defined as $\inf_{y \in Y} d(x,y)$ and $d(x,y)$ is computed with Euclidean distance in our experiments. 

To comprehensively evaluate the results, we also use average symmetric surface distance (ASSD), which quantifies the average distance between the surfaces of two segmented objects. Compared to Hausdorff distance, it measures how well the boundary of the segmented object and the ground truth align with each other on average. It is defined as
\begin{align}
\textrm{ASSD} = \frac{ \sum_{x\in X} d(x, Y) + \sum_{y\in Y} d(X, y) }{|X| + |Y|}
\end{align}
where $|X|$ means the length of the contour $X$.

\subsection{Comparative study}
To evaluate the performance of our proposed model, we first compared our model without transfer learning with other commonly used segmentation models. The average performance of each model on the CCF dataset is shown in Table~\ref{compareccf}. We observe that our model outperforms other models in terms of precision, Dice score, IoU and average symmetric surface distance. While the proposed model achieves a relatively low sensitivity, our precision value is significantly higher than others and therefore gets the best overall performance. 

\begin{figure*}[ht]
\centering
\includegraphics[width=0.9\textwidth]{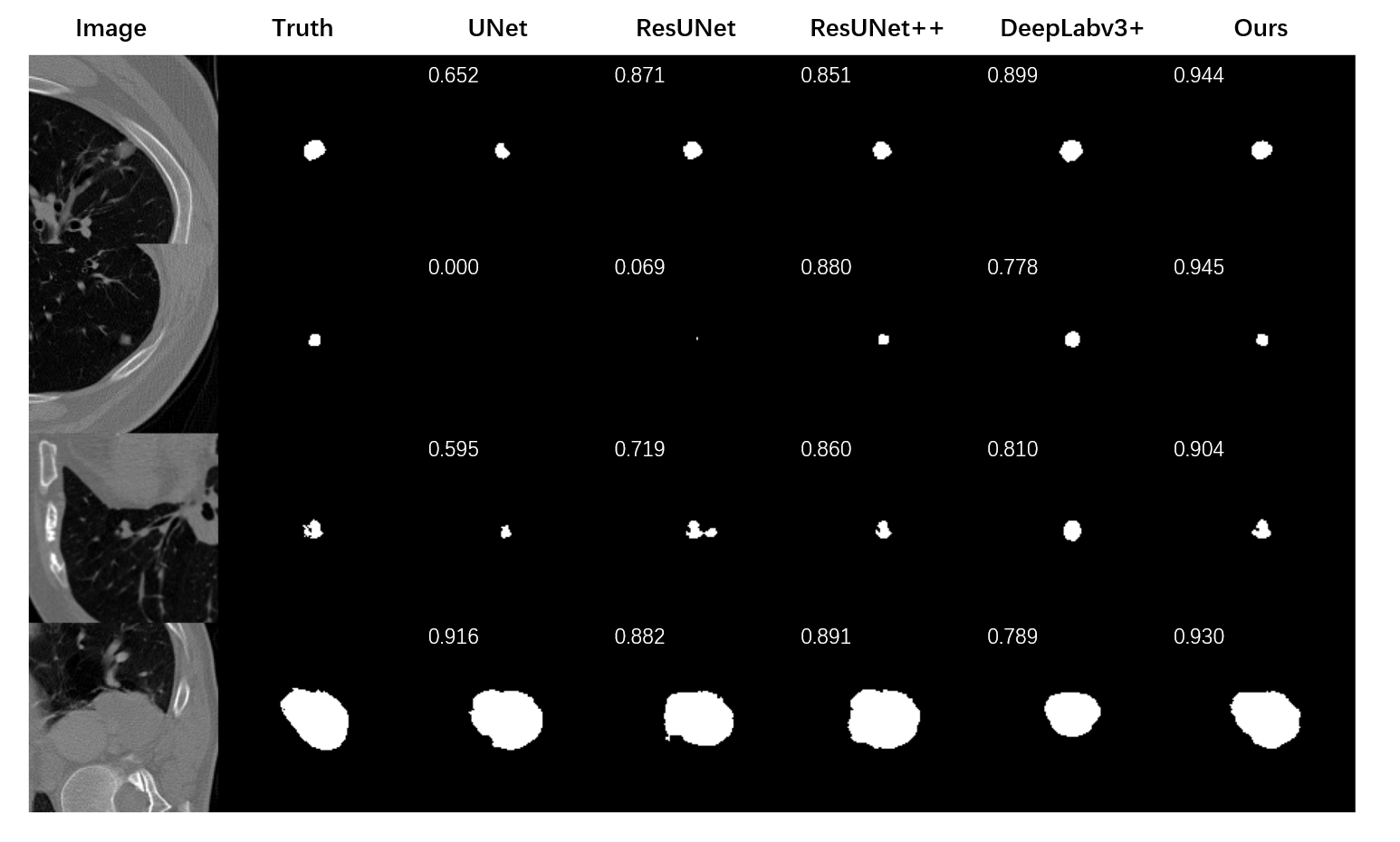}
\caption{Visual comparison of segmentation results for different methods, including four examples in the CCF dataset. The dice score of each result is shown on the upper left of each prediction.}
\label{figure4}
\end{figure*}

We also visualized the segmentation results for comparison. We demonstrate four examples of the CCF dataset from different patients in Figure~\ref{figure4} with the prediction of the compared models and corresponding dice scores of them. The first example is a typical juxtapleural nodule with some heterogeneous attenuation within it. Although all the five models successfully identified it, UNet, ResUNet and ResUNet++ just captured the parts with relatively higher intensity and resulted in lower Dice scores. In the second example, UNet and ResUNet failed to identify the tiny nodule, while our model segmented the nodule accurately. In the third and fourth examples, our model captured the shapes of the nodules accurately resulting in the best Dice score among the models.

\begin{table*}[width=1.9\linewidth,cols=7,pos=h]
\caption{Comparative study with CCF dataset.}
    \begin{tabular*}{\tblwidth}{@{} LLLLLLL@{} }
         \toprule
         Model & Pre & Sen & Dice & IoU & HD(mm) & ASSD(mm) \\
         \hline
         UNet \cite{unet} \cite{} & 0.832 & 0.821 & 0.753 & 0.688 & 4.591 & 0.341 \\
         ResUNet \cite{resunet} & 0.821 & \textbf{0.855} & 0.779 & 0.708 & 3.594 & 0.290 \\
         ResUnet++ \cite{resunetplus} & 0.844 & 0.845 & 0.777 & 0.712 & 3.784 & 0.303 \\
         DeepLabv3+ \cite{deeplabv3plus} & 0.802 & 0.831 & 0.748 & 0.658 & 3.318 & 0.265 \\
         Proposed Model & \textbf{0.870} & 0.827 & \textbf{0.801} & \textbf{0.735} & \textbf{2.929} & \textbf{0.211} \\
         \bottomrule
    \end{tabular*}
    \label{compareccf}
\end{table*}

\subsection{Experiments for transfer learning strategies}
During pre-training, the neural network model learns useful feature representations from the large datasets. Therefore, the parameters of some layers are usually frozen during the training process to prevent them from being updated based on the gradients computed from the loss function. By freezing some layers, the learned features are preserved and not overwritten during the fine-tuning process. Additionally, freezing layers can reduce computational cost and the risk of overfitting since there are fewer parameters to update. The choice of which layers to freeze depends on factors such as the similarity between the small and large datasets. In an encoder-decoder based model, the early convolutional layers in encoder block are typically responsible for extracting low-level features like edges and textures, which are often generalizable across different datasets. These layers can be frozen if the pre-trained weights are from a dataset similar to the target training data. The final layers in encoder block can capture high level and task-specific features, and the decoder layers are responsible for upsampling and combining features from different resolutions.

To ensure the best result using transfer learning, we conducted a set of experiments to determine the optimal transfer learning strategy, as shown in Table~\ref{tl}. We froze S1 (the first segmentation encoder block), S5 (the last segmentation encoder block), S9 (the last segmentation decoder block), and C1 (the first classification block). Since both the pre-training and the fine-tuning involve binary classification, we also tested the case when freezing FC (the fully connected layer of the classification part). The results in Table~\ref{tl} indicate that fully fine-tuning without freezing any layers outperforms all the other strategies, demonstrating that there are essential differences between LUNA16 and the CCF dataset. However, the ablation study in the next section shows that the model still benefits from pre-training on LUNA16.
\begin{table*}[width=1.9\linewidth,cols=7,pos=h]
\caption{Experiments testing different TL strategies}
    \begin{tabular*}{\tblwidth}{@{} LLLLLLL@{} }
         \toprule
         Model & Pre & Sen & Dice & IoU & HD(mm) & ASSD(mm) \\
         \hline
         W/o freezing & 0.828 & 0.885 & \textbf{0.814} & \textbf{0.744} & \textbf{3.188} & \textbf{0.280} \\
         Freeze S1 & 0.804 & \textbf{0.889} & 0.798 & 0.728 & 3.611 & 0.310 \\
         Freeze S5 & 0.817 & 0.888 & 0.810 & 0.741 & 3.608 & 0.296 \\
         Freeze S9 & \textbf{0.879} & 0.709 & 0.708 & 0.627 & 4.644 & 0.401 \\
         Freeze C1 & 0.824 & 0.879 & 0.802 & 0.734 & 3.394 & 0.287 \\
         Freeze FC & 0.815 & 0.884 & 0.799 & 0.728 & 3.226 & 0.295 \\
         \bottomrule
    \end{tabular*}
    \label{tl}
\end{table*}

\subsection{Ablation study}
We conducted an ablation study to assess the effect of each component in our model, as presented in Table~\ref{ablation}. In the first row, we implemented the proposed backbone without feature combination blocks and STD regularization. This means the segmentation part works independently without being combined with classification. The results show a Dice score of 0.755, Hausdorff distance of 3.509 mm, and average symmetric surface distance of 0.276 mm. In the second row, we implemented the proposed model without STD regularization and observed better performance in all metrics except average symmetric surface distance. This highlights the improvement brought by the feature combination blocks. In the third row, we applied STD regularization into our model but setting the value of $\lambda_2$ in Equation~\ref{newstd} as zero to show the effect of the third term in Equation~\ref{newstd}. Comparing the second and third rows, it is evident that STD regularization leads to a noticeable improvement in precision and Dice score. In the fourth row, we implemented the model to show the effect of the fourth term in Equation~\ref{newstd}. Compared with the second the third rows, all the metrics except sensitivity are significantly improved. Finally, we applied the transfer learning strategy in our model, achieving a Dice score of 0.814 and IoU of 0.744. Although transfer learning leads to slightly worse HD and ASDD, it again shows the difference between the two datasets and that the model still benefits from transfer learning.

\begin{table*}[width=1.9\linewidth,cols=7,pos=h]
\caption{Ablation studies.}
    \begin{tabular*}{\tblwidth}{@{} LLLLLLL@{} }
         \toprule
         Model & Pre & Sen & Dice & IoU & HD(mm) & ASSD(mm) \\
         \hline
         Backbone w/o combination & 0.820 & 0.848 & 0.755 & 0.693 & 3.509 & 0.276 \\
         Backbone w/o STD & 0.823 & 0.849 & 0.764 & 0.704 & 4.020 & 0.285 \\
         Backbone + STD with $\lambda_2=0$ & 0.846 & 0.847 & 0.787 & 0.720 & 3.927 & 0.287 \\
         Backbone + STD & \textbf{0.870} & 0.827 & 0.801 & 0.735 & \textbf{2.929} & \textbf{0.211} \\
         Backbone + TL + STD & 0.828 & \textbf{0.885} & \textbf{0.814} & \textbf{0.744} & 3.188 & 0.280 \\
         \bottomrule
    \end{tabular*}
    \label{ablation}
\end{table*}

\section{Conclusions}
In this paper, we present a multitask deep learning model designed and implemented for segmenting small clinical lung nodule datasets. Our model integrates segmentation and classification components through feature combination blocks to facilitate information sharing between the two tasks. We incorporate spatial regularization adaptively guided by the classification output in the segmentation process. Additionally, we employ transfer learning by pre-training the model on LUNA16, a large public dataset. Our experimental results demonstrate that the proposed model surpasses other classic models in the clinical dataset. We conduct a series of experiments by freezing different hidden layers during fine-tuning to determine the optimal transfer learning strategy. Furthermore, our ablation study showcases the enhancements provided by various components within our model.

While our model demonstrates promising results, there remains ample room for improvement, particularly when applied to small clinical datasets. Our model, especially the classification component, is designed to be simple and efficient for such datasets. However, we recognize that there is considerable flexibility in altering both the classification and segmentation components. Additionally, the distribution of feature combination blocks may need to be modified to further enhance the model's performance on specific datasets.

\section*{Credit author statement}
\textbf{Jiasen Zhang:} implementation, conceptualization, methodology, evaluation, writing.
\textbf{Mingrui Yang:} methodology, validation, supervision, review \& editing.
\textbf{Weihong Guo:} methodology, conceptualization, supervision, review \& editing. 
\textbf{Brian A. Xavier:} clinical data, annotation, conceptualization.
\textbf{Michael Bolen:} clinical data, annotation, conceptualization.
\textbf{Xiaojuan Li:} investigation, supervision, review \& editing.

\section*{Conflict of interest}
The authors declare no conflict of interest.

\section*{Acknowledgements}
This work was supported by the Cleveland Clinic Research Co-Laboratories Award.

\printcredits

\bibliographystyle{cas-model2-names}

\bibliography{cas-refs}




\end{document}